\documentclass[a4paper]{article}
\usepackage[utf8]{inputenc}
\usepackage[T1]{fontenc}
\usepackage{graphicx}
\usepackage{longtable}
\usepackage{wrapfig}
\usepackage{rotating}
\usepackage[normalem]{ulem}
\usepackage{amsmath}
\usepackage{amssymb}
\usepackage{capt-of}
\usepackage{hyperref}
\usepackage{longtable}
\usepackage{pdflscape}
\usepackage{parskip}
\usepackage[table]{xcolor}

\usepackage{authblk}

\title{Large Language Models and Emergence: \\ A Complex Systems Perspective}
\author[1] {David C. Krakauer}
\author[1,2] {John W. Krakauer}
\author[1] {Melanie Mitchell}
\affil[1]{Santa Fe Institute}
\affil[2]{ Department of Neuroscience, Johns Hopkins School of Medicine}
\date{\today}

\begin{document}
\maketitle

\begin{abstract}
\noindent \textit{Emergence} is a concept in complexity science that describes how many-body systems manifest novel higher-level properties, properties that can be described by replacing high-dimensional mechanisms with lower-dimensional effective variables and theories. This is captured by the idea ``more is different''. Intelligence is a consummate emergent property manifesting increasingly efficient---cheaper and faster---uses of emergent capabilities to solve problems. This is captured by the idea ``less is more''.  In this paper, we first examine claims that Large Language Models exhibit emergent capabilities, reviewing several approaches to quantifying emergence, and secondly ask whether LLMs possess emergent intelligence. 
\end{abstract}

\section{Introduction}

Large Language Models (LLMs) are deep neural networks that, through training on huge amounts of text, learn to accurately predict the next word (or token) in a text.  It has been surprising to many that next-token prediction has lead to impressive abilities, such as learning of syntax, code generation, writing in any style, and factual recall. It has been claimed in the LLM literature that, as the number of network parameters and amount of training data is scaled up, certain capabilities arise suddenly and unexpectedly, a phenomenon that these writers term ``emergence''.  For example, Wei et al.\ \cite{wei2022emergent} write, ``we define emergent abilities of large language models as abilities that are not present in smaller-scale models but are present in large-scale models; thus they cannot be predicted by simply extrapolating the performance improvements on smaller-scale models.'' And in a recent review of emergent abilities in LLMs Berti et al. \cite{berti2025emergent} survey around 100 papers the majority of which equate emergence with the discontinuous appearance of abilities with increasing data or model size. 

Several questions are immediately raised by such statements. First, what does emergence mean? Second, why does emergence matter? And finally what is meant by ``cannot be predicted''?  In the scientific literature, emergence is a term with a more rigorous underpinning than merely colloquial reference to either the discontinuous development of capability, or the unexpectedness of capacities from the point of view of a human observer. Minimally, emergence describes the reorganization of a system that can support a new, often smaller description, that screens off microscopic details not essential to predicting the future of a system.  Emergence matters because it leads to an enormous cost saving in how systems are described, predicted, and controlled. We do not need to use quantum mechanics to build a bridge because the classical world emerges from the quantum---a fact exploited by engineers.  We similarly do not need to track every individual in an economy to predict economic recessions: macro-level variables such as stock prices, unemployment levels, and interest rates help provide effective (though imperfect) economic theories.



In this paper we argue that the term emergence, in the context of LLMs, should be reserved for the combination of successful task performance and the associated new coarse-grained representations formed within the responsible structure, in this case the neural network. Evidence for emergence could include  novel bases that promote algorithmic compression and new internal models of computation,  abstractions that lead to demonstrable efficiencies in prediction, problem solving and generalization, and the breaking of scaling -- breaking self-similarity -- with increasing data and model size.  Additionally, in the context of LLMs, we argue for the importance of distinguishing emergent \textit{capabilities} from emergent \textit{intelligence}. 

Emergent intelligence is the internal use of coarse-grainings to solve a broad range of problems. In humans these coarse-grainings are often related through analogy, that by means of thoughtful and minimal modification, permit significant generalization and extrapolation. For example, the inverse square law is a coarse-grained concept, that, through analogies made in the physical sciences, describes behavior  relating to gravity, electrostatics, acoustics, and electromagnetism.

By contrast, calculators are capable, they have many engineered (rather then emergent) functions that encode a large number of efficient mechanisms for solving problems of arithmetic, including the Taylor expansion, Sterling's formula, and the CORDIC algorithm. We would not call these intelligent because they cannot support the construction of analogies between these concepts, and provide no mechanism for the simple modification of rules into new capabilities.

Emergent intelligence is clearly a feature of human reasoning and human language, but as of yet, an unproven feature of LLMs, which at best, as we shall argue below, demonstrate emergent capability. Many of these capabilities  are likely to  exceed human capability the ways calculators have been exceeding human arithmetical ability for over a century. To be fair LLMs, have never been selected for efficiency through evolution, and in some cases have been shown to exhibit a Rube Goldberg logic, whose most consistent feature as described by Anthropic Research, ``is the massive complexity underlying the model’s responses even in relatively simple contexts. The mechanisms of the model can apparently only be faithfully described using an overwhelmingly large causal graph'' \cite{lindsey2025biology}.  And this is coupled to extremely inefficient energy use \cite{jiang2024preventing}. 

There is little reason to expect LLMS to be intelligent since all we have been training through endless benchmark targeting is hugely overparameterized capability. A gifted mathematician is clearly not just a vast assemblage of diverse calculators; they are much closer to an analogy-making system, typically in possession of rather poor calculators. A mathematician is described as intelligent because they can do ``more with less'' (explaining an ever increasing number of phenomena with a modest set of basic ideas, or as Henri Poincare wrote, "Mathematics is the art of giving the same name to different things.") not ``more with more'' (explaining an ever increasing number of phenomena with an ever increasing set of contingent ideas).  

In this paper, we begin by describing the specific claims in the LLM literature concerning the emergence of capabilities.  We then provide a short ``primer'' on the concept of emergence, and the conditions under which we believe it makes sense to use the term.  We contrast two types of emergence, which we term ``knowledge-out'' (properties emerging from systems with simple homogeneous components) and ``knowledge-in'' (properties emerging from systems with complex structure or with complex inputs or environment), and we disucss the importance of this distinction.  Following this, we lay out a framework for analyzing emergence using mechanisms studied in complexity science, including scaling, criticality, compression, and novel bases and manifolds, and we describe how emergent coarse-graining of a system can enable generalization abilities.  We conclude by discussing the important difference between systems displaying emergent capabilities and those displaying emergent \textit{intelligence}.

\section{Emergence Claims for LLMs}
The 2022 paper ``Emergent Abilities of Large Language Models'' \cite{wei2022emergent} introduced the idea that as network and training data size are scaled up, LLMs exhibit sudden, unexpected ``emergent'' capacities that were not present in smaller models.  The authors plot model accuracy versus scale for LLMs of various sizes on several benchmark datasets. For many tasks, at certain scales each model shows a sharp increase in accuracy, an increase that was not predicted by earlier empirical ``scaling laws'' \cite{kaplan2020scaling}. In some cases the accuracy remains at or near zero over several scales until it sharply rises well above random chance. One example is a benchmark that tests accuracy on three-digit addition.  As described in \cite{berti2025emergent}, a model with six billion parameters achieves only 1\% accuracy on this benchmark, a model with 13 billion parameters improves only slightly to 8\% accuracy, but a model with 175 billion parameters suddenly reaches 80\% accuracy.  Wei et al.\ note that ``emergence motivates future research on why such abilities are acquired and whether more scaling will lead to further emergent abilities.''

There has been some controversy over how sharply these improvements occur as networks are scaled; a follow-up study  \cite{schaeffer2023emergent} showed that for some of the capabilities termed ``emergent'' in \cite{wei2022emergent}, if more continuous success metrics are used, the improvement with scale becomes continuous rather than abrupt.  However, other studies have questioned the generality of the continuous improvement claims \cite{wei2022emergent,berti2025emergent}.

It has also been argued that so-called emergent abilities of LLMs result from a (predictable) increase in ability in larger models for ``in-context learning'', that is, an improved ability to use examples given in a prompt, as well as from post-training ``instruction tuning'', which improves the ability of larger models to follow instructions given in a prompt \cite{lu2023emergent}. 

Other studies claiming emergence in LLMs have equated emergent abilities simply with those abilities that LLMs were not explicitly trained for, such as numerical understanding \cite{noever2023numeracy}, analogical reasoning \cite{webb2023emergent}, capability for legal reasoning \cite{nay2024large}, and formation of ``world models'' \cite{li2023emergent}.

In summary, in the LLM literature, the term \textit{emergence} has been used to refer to surprising or sudden jumps in accuracy on particular benchmarks as a function of the scale of the data, model, or cluster, or as capacities of models that they were not explicitly trained to have.  (A similar analysis of emergence claims for LLMs is given in \cite{rogers2023position}.)

In the following sections, we will describe a richer and more useful framework for understanding emergence, one that has a long history in the scientific literature. Very few of the features of LLMs, from the abruptness of performance increases on benchmarks, through to generalization, have much, if anything to do with any technical sense of the word emergence, and are adequately described using the more familiar ideas of learning, inference, compression, and perhaps development.  Here, we situate claims of emergent abilities---and more generally, emergent \textit{intelligence}---in LLMs in the framework of complexity science.  



\section {A Primer on Emergence \label{sec:primer}}

There is no single accepted definition of emergence \cite{Bedau2008-qn,Gibb2019-lp}. However, there are widely accepted necessary (though not sufficient) mereological properties (part-to-whole relations). Emergence necessarily permits a new and compressed model or description of a system's long-term behavior: emergent properties are not efficiently described at their microscopic levels but through coarse-graining, allowing for ``effective theories'' operating on derived degrees of freedom \cite{Chalmers2006-in, krakauer2020information, Carroll2024-mx}. 

For example, populations of molecules are often described using individual-level molecular dynamics. But they can also be described parsimoniously through their aggregate properties by treating them as a continuum using fluid dynamics \cite{blazek2015computational}.  Molecular dynamics needs to account for each particle's position and momentum independently, whereas fluid dynamics need only describe bulk properties: total mass, length, pressure, and time. In this way, fluid dynamics provides an effective theory of these coarse-grained variables---the emergent properties of fluids. Such an effective theory can describe and predict a system at the coarse-grained level without any reference to the lower level. There is no requirement for discontinuity, or ``abrupt changes'' in the phenomena, merely a new description that efficiently captures the new organization. 

In short, a minimum and necessary signature of any emergence claim is a coarse-graining of observables coupled to a compression of the system description that remains predictive \cite{israeli2006coarse}. This is what is implied by Anderson's famous ``more is different'' maxim \cite{Anderson1972-pw}. The word ``different'' here refers to the novel, coarse-grained variables and macroscopic rules describing the time evolution of any system, which enables the effective theory to screen off contributions from microscopic degrees of freedom \cite{salmon2010statistical}. Anderson was specifically interested in how the symmetric dynamics of physical laws cease to be informative with increasing molecular scales, where symmetry is often broken, and how the loss of configurational symmetry demands new coarse grained models \cite{krakauer2023symmetry}. This basic argument for the power of "effective theory" is lost in almost all discussions of emergence in the LLM literature, as is illustrated in Berti et. al's recent review \cite{berti2025emergent}.


Emergent properties are typically observed following the addition of constituent elements (e.g., adding molecules to a medium or adding edges to a graph) or through changes in the internal organization of a system (increasing temperature or rewiring existing edges) \cite{Krakauer2024-vu}. A system is emergent when a new and causally sufficient reduced description of a system is enabled by these modifications. 



Most generally, a central aspect of emergence is the appearance of novel or compressible structure and function. A second aspect of emergence is that this structure and function can come ``for free'', referring to self-organization, or more fundamental physical and chemical processes, which produce ordered states through a process of energy minimization.  We can say that the term emergence typically applies when at least some of the following conditions hold:

\subsection*{Conditions for Emergence } \label{list}

\begin{enumerate}
\item \textit{Scaling:} A new internal organization arises as a function of scaling of generic components that lends itself to a new and parsimonious description.
\item \textit{Criticality:} Systems that undergo a rapid or discontinuous change, such as a phase transition in a control variable, where a new organization lends itself to a new description.
\item \textit{Compression:} Compressed representations  internal to the model are exploited by the model to increase the representation's fidelity or efficiency.
\item \textit{Novel Bases:} New bases or functions are discovered that provide an internal ``alphabet'' that is used to encode regularities. 
\item \textit{Generalization:} In adaptive systems the capabilities arising from one task can be used to solve different tasks, enabling competence arising from the union of discrete performance. 
\end{enumerate}

This list makes clear why emergence is so important. Systems that demonstrate emergent properties can be induced to produce novelty through a simple scaling of parts and they lend themselves to very efficient, high-level descriptions or abstractions, which provide levers for control and design. Without emergence, the only way to make sense of and influence the world would be through microscopic descriptions of its basic units and interactions, that is, via a scientific monoculture described using quantum field theory with interventions at the most fundamental scales. 

There is a complication for identifying emergence in software: all engineered systems need to possess causally emergent capabilities because they need to screen-off irrelevant microscopic details. However, engineered systems are rarely described as ``emergent'' because they arise by design---they are designed or trained to produce a target macroscopic behavior. This in contrast to, for example, the pizoelectic properties of a quartz crystal that arise when sufficient collections of oxygen and silicon atoms are bonded at high pressure. The macroscopic pizoelectric property was never a target of design, learning, or selection; it is a collective property of atomic interactions. In order to discuss whether LLMs show emergence, given that they are, in a sense, engineered, we need to make a distinction between two different kinds of emergence in complex systems, which we term "knowledge-out" and "knowledge-in", and describe in the next section. 

\section{Knowledge-Out Versus Knowledge-In Emergence}

 There is a useful distinction to be made between the kind of emergence seen in many-body physical and chemical systems and that seen in complex \textit{adaptive} systems.  In the former case, emergent properties are assumed to be generated from large numbers of simple components following relatively simple rules (e.g., individual molecules obeying Newton's laws). In contrast, in complex systems that adapt or learn, such as biological organisms, brains, economies, and LLMs, emergent properties such as bodily organs, stock indices, or cognitive capabilities are generated by mechanisms that extract large amounts of ``knowledge''---structured information including facts and rules---from pre-existing complex environments.  We will say that emergence in chemistry and physics has a ``knowledge-out'' (KO) character, since it is characterized by complex structure or behavior arising from simple interactions among simple elements. Conversely, emergence in complex adaptive systems has a ``knowledge-in'' (KI) character, since it is characterized by complex structure or behavior arising from complex inputs or environments \cite{Krakauer2024-vu}.  

The objective of KO science is to demonstrate how parsimonious and local mechanisms, when implemented by large collections of identical (or near identical) elements, produce macroscopic outputs that permit a variety of knowledge-rich explanations \cite{Feynman1963-jv}. When it comes to the KI sciences, emergence claims are not so straightforward, given that  adaptive changes are made to every component in a system on an individual basis. 

In KI cases, the word emergence is often substituted with the words engineered, developed, evolved, trained, or learned. In reviewing the use of emergence concepts from physics versus biology, Newman refers to these distinctions as ``generic versus genetic'' \cite{Newman2015-dk}, which correspond to KO and KI, respectively. In a \textit{genetic} or KI system, where each component is parameterized, there is no obvious ``free'' structure and function of the kind found in \textit{generic} or KO systems such as collections of atoms. However, in KI systems, there can still be coarse-grained global properties that screen off microscopic details. Given that LLMS's are clearly KI systems---trained with vast corpora using machine learning methods, emergence claims are required to describe both their coarse-grained global properties as well as the local microscopic mechanisms from which they emerge. 

In Table \ref{tab:t1} we illustrate this micro-macro requirement by listing a few examples of phenomena that are widely thought to exhibit KI-type emergence in evolved and engineered systems.  In each example, the local mechanism is a property or action of an individual in a population (e.g., an insect in a colony, an individual activator or inhibitor in a medium, a bird in a flock, a neuron in a population), each of which is the product of evolutionary processes in complex environments. The local, microscopic interactions enable global macroscopic properties, which enable coarse-grained descriptions that are themselves the elements of an effective theory, one that is casually sufficient to predict the long-term behavior of the system.  In none of the examples can emergence be claimed based purely on macroscopic properties.

\begin{table}
\small
    \centering
    \begin{tabular} {|p{3cm}|p{3cm}|p{3cm}|l|}\hline 
         \textbf{Emergent Phenomenon}&  \textbf{Local Microscopic Mechanism}& \textbf{Global Macroscopic Property}&\textbf{Reference}\\ \hline 
         Stigmergy&  Depositing an activating chemical trace&  Efficient resource allocation/extraction & \cite{Heylighen2016-be}{}\\ \hline 
         Turing Patterns&  Differential diffusion of activators and inhibitors&  Stripes, segments, and spirals& \cite{Turing1952-dd}\\ \hline 
         Flocking&  Separation, alignment, and cohesion&  Coordinated object avoidance& \cite{Reynolds1987-fh}\\ \hline 
         Visual Receptive Fields&  Maximizing mutual information between cells&  Center surround filters& \cite{Linsker1990-qh}\\ \hline 
         Content-Addressable Memory&  Asynchronously spiking thresholded units&  Error correcting memory and generalization& \cite{Hopfield1982-ph}\\ \hline
    \end{tabular}
    \caption{A few examples of KI-type emergent phenomena including local mechanisms, which are either evolved or learned, and coarse-grained global properties.}
    \label{tab:t1}
\end{table}

\section{Framework for the Analysis of Emergence}

What kind of emergence is demonstrated by an LLM? What is the relevant coarse-graining and compression and what consequent effective theories for their behavior provide evidence of emergence? \cite{krakauer2023unifying}.  In this section we attempt to situate LLMs in the five principles that we sketched in Section~\ref{sec:primer}: (1) Scaling: how changing the number of system components changes its properties; (2) Criticality: the theory of the phases of a system; (3) Compression: reduction of the system's description size or dimensionality achieved through efficient coarse-graining; (4) Novel Bases: discovering the minimal constituent elements capable of describing the system; and (5) Generalization: the behavior of system rules outside of the training or adaptive context. Most emergence claims in the scientific literature make use of some combination of these principles. We also relate these claims to the challenge of describing emergence in the KI setting.  

\subsection{Emergence Through Scaling and Criticality}

Scaling is often applied to emergence in complex systems to demonstrate how ``more is different''. More components can support more efficient descriptions (as with the previous example of fluid dynamics) as well as supporting phase transitions (critical behavior). By changing the value of a control variable (component number, temperature, etc) there can be a corresponding causal change in an order parameter (e.g., magnetization) that shows power law scaling in the vicinity of a phase transition \cite{Sole2011-rt}.  Scaling in system size (size of molecules, connections,etc) can lead to the spontaneous breaking of a symmetry, such that different phases are described by different symmetry groups \cite{flack2013timescales, schuster2014criticality, o2015backbones}. These emergent regularities are often captured through effective parameters that describe the emergent dimensions in a model. 

This ``more is different'' scaling is unlike the idea of scale-invariant properties of a system \cite{brown2000scaling} or the scaling of distributions of data  \cite{Thurner2018-gk}.  These allometric and distributional scaling principles become important as a reference or baseline against which to capture fundamental changes in internal organization through the ``breaking'' of scaling---different exponents are required to capture regularity at different system sizes where this scale is not set externally \cite{kempes2019scales}.

The double descent phenomenon seen in neural networks \cite{nakkiran2021deep}, in which, as model size increases, test loss goes down, then up due to overfitting, and then down again, has been proposed as an example of emergence through the breaking of scaling. However, both linear and polynomial regression demonstrate double descent under the appropriate conditions. These relate to the choice of penalty and or complexity measures on either side of an interpolation threshold \cite{curth2023u}, and how representative samples are of the dimensions of variation in the test set \cite{schaeffer2023double}. Thus double descent provide no evidence for emergence. 

More persuasive is recent work by Guth \& Ménard (2025, in prep) showing that for neural networks being trained on a given task, the peak of the double descent behavior in the loss occurs concomitantly with a qualitative change in the neural encoding: the covariance spectra of weights (or learned features) go from exponential to scale-free. This qualitative reorganization implies a qualitative change in the bases available to perform the task. This would seem to provide evidence  for emergence in a convolutional neural network in the form of a phase transition giving rise to a new form of representational structure inside the network associated with an improvement in performance. This is an emergent capability.

Scaling has come to play a leading role in claims of emergence in LLMs.  Recall that a major claim is that scaling up network parameters or training data results in discontinuous jumps in accuracy on certain benchmarks---jumps that roughly resemble phase transitions in physical systems (such as water's phase transitions under increasing temperature) \cite{chen2023sudden} \cite{nakaishi2024critical}.  However, there are many differences between such discontinuities and physical phase transitions. First, it is not clear that LLM abilities have clear ``phases'', or if the observed jumps in accuracy are actually continuous improvements under certain metrics \cite{schaeffer2023emergent}. Second, in physical phase transitions, the control parameter (e.g., temperature) is typically a one-dimensional quantity.  While LLM capability transitions are plotted on a two-dimensional graph, the control variable on the x-axis (``scale'') is actually a high-dimensional complex quantity---either human-generated text data consisting of highly interdependent tokens or the highly interdependent, evolving parameters of a network-in-training, or both at the same time. This is what we have described as Knowledge-In (KI), and while KI does not rule out emergence (see Table~\ref{tab:t1}), it does require that we distinguish carefully between the local mechanisms and the global property of interest. If the global property is being ``programmed'' by extensive training, we would hardly describe the behavior as emergent.   

Furthermore it is not clear that discrete jumps in LLM accuracy (if such jumps exist) reflect any kind of global reorganization of the internal structure of the network that supports the coarse-graining required by the most elementary notion of emergence. As we have argued in making the KO and KI distinction, in the KI case evidence from  the external behavior of LLMs is not sufficient to establish that the network has transitioned to a new ``phase'' with novel emergent capabilities; it will be necessary to better understand what these behaviors correspond to microscopically ``under the hood''.  A candidate study supporting emergent capability through scaling is \cite{chen2023sudden},  which provides evidence for a non-linear reduction in a test loss concomitant with the acquisition of a new internal syntactic structure, Where this structure also seems to support a reduced-complexity description. This is rather like the flocking example \cite{Reynolds1987-fh} in which, given sufficient agents (in the LLM case, the size of the networks and training data), new  collective behavior emerges with low dimensional degrees of freedom. 

\subsection{Emergence Through Compression}

Compression describes a multitude of approaches for encoding information or mechanisms in a lower-dimensional representation. This is typically achieved by eliminating redundancies, efficiently encoding common features, and quantizing sparse data.  Emergence is associated with the internal discovery of coarse-grained models that capture the compressed regularities in the data. The strongest evidence of emergence through compression is when the system operates directly on the coarse-grained degrees of freedom in order to perform better inference, such as in sensory perception \cite{olshausen2004sparse, stevens2015fly}.  

A possible example of emergence through compression in LLMs is the case of ``emergent world models'' in the OthelloGPT transformer \cite{li2023emergent,nanda2023emergent}. In this case, a transformer trained only on tokens representing legal moves in the game Othello was shown to have formed compressed, causally relevant internal representations of the Othello game board---in other words, a coarse-grained model that captures the compressed regularities in the data.  This is a different form of emergence than that claimed in \cite{wei2022emergent}, in which new abilities seemed to arise with increased scale.  In OthelloGPT, no scaling was involved; the term ``emergence'' was used to refer to the creation of a coarse-grained, parsimonious internal model of the data.  One caveat is that the internal model may not have been as parsimonious or ``compressed'' as implied in the original papers---another study claimed that the internal model amounted to a large ``bag of heuristics'' \cite{lin2024bag}. Moreover, there was only weak evidence that the emergent representation was causally responsible for the network's ability to predict legal moves, and so at this point we can not be certain that this is a truly emergent capability.

\subsection{Emergence Through Novel Bases and Manifolds}

An important approach to compression is to discover novel bases: encodings of a subspace in a dataset than can jointly span its full space. Bases can be composed in principled ways to recover a full data set. Unlike simple forms of data compression (lossy or lossless), bases are compositional and interpretable. In this case emergence takes the forms of the appearance of a small number of novel bases to describe what would otherwise be a very large dataset \cite{Linsker1990-qh, kaschube2008self}. 

Manifolds---spaces that are locally Euclidean and whose bases are open sets---provide a natural means of connecting points that live in high dimensional spaces. The manifold hypothesis asserts that high dimensional data sets can be encoded without loss of information, through local low dimensional neighborhoods. The emergence of a manifold (e.g. color space, frequency space) is thought of as an efficient means of encoding regularities as well as interpolating and extrapolating to new points \cite{polani2008foundations, yin2008learning, rupe2024principles}.  

As in emergence through compression, establishing emergence through novel bases and manifolds requires understanding how the internals of a system have organized to implement such structures.  Relatively little is known about whether or how such structures might exist in LLMs, beyond the kinds of simple ``mechanistic interpretability'' analyses such as that described above for OthelloGPT.  Two of the more compelling examples of novel bases include the demonstration of abstraction units in self-supervised transformers \cite{ferry2023emergence} and the discovery of covariant features in visual inputs that are aligned through layers of geometric transformations into very regular patterns of activation \cite{guth2024rainbow}. 

\subsection{ Generalization Through Emergence}
We use the term ``generalization'' to refer to the ability of an adaptive system to succeed on tasks or in situations that differ qualitatively from the system's prior experiences or training.  Emergence---in the form of coarse-grained variables and theories---can give rise to robust generalization by transforming a detailed, high-dimensional representation to a more abstract, low-dimensional one, via compression and the discovery of novel bases. For example, evolution produced groups of neurons in the brain that become organized as spatial maps, structures that represent higher-level units of information than individual neurons, and that generalize across myriad situations, including being used to process non-spatial data. Indeed, this kind of ability for generalization is what gives emergence much of its power.  

Generalization has been approached formally in several ways, including through measures of dimension, (e.g, the Vapnik–Chervonenkis (VC) dimension, which provide a natural measure of the ability of a system to classify points \cite{vapnik1994measuring}), and through measures of function approximation (e.g. The Probably Approximately Correct framework, which provides a measure of the generalization gap \cite{valiant2013probably}). Both of these provide techniques for analyzing generalization in stationary environments, whereas emergence in humans often implies behavior in radically different environments. 


Given that the training data and the task-specific internal mechanisms for LLMs are typically unknown, it is hard to determine whether the claimed abilities are general, or whether they rely on memorization of similar examples in the training data or on non-generalizable ``shortcuts'' \cite{mitchell2023we}.  In a number of cases, LLM abilities cited as emergent have later been shown to lack generality and robustness (e.g., \cite{mirzadeh2024gsm,lewis2024evaluating}). Nevertheless, there is some evidence for such generalization ability in LLMs, for example,  extrapolating forward in time the behavior of real-valued dynamical systems, without fine tuning when using language  data to train the networks. This opens up a series of fascinating questions about the domain-generality of  rules encoded in natural language \cite{liu2024llms}. 

\section{Conclusions}

\subsection{Emergent Capabilities in LLMs}

In this paper we surveyed some of the claims for emergent capabilities in LLMs.  As background for evaluating these claims, we described necessary conditions for emergence, as well as a framework of several principles for quantifying it.

We argued that in LLMs, the term emergence should be used not merely to signify surprising or unpredictable task performance, or abrupt changes in performance, but requires at minimum the identification of relevant coarse-grained variables that form effective mechanisms--- reduced ``internal degrees of freedom''---for this behavior, mechanisms that can explain or predict the behavior of the system at this higher level, screening off details of lower level mechanisms such as weights and activations.  More quantitative evidence for emergence includes the kinds of principles related to emergence in physical systems, such as breaking of scaling through reorganization, evidence for the use of novel bases and manifolds formed through compression of regularities, and new forms of abstraction that lead to demonstrable efficiencies in prediction, problem solving,  generalization, and analogy-making.  Identifying such principles would be an important step in understanding the seemingly novel capabilities that arise in LLMs.  

Three types of emergence claims have been made for LLM capabilities: (1) sharp improvements in specific capabilities that occur as the system or training data is scaled; (2) capabilities are identified that the LLMs were not specifically trained for; and (3) internal ``world models'' emerging from autoregressive token prediction.  Each of these cases, and particularly the last, present provocative evidence for emergence, but in all cases that evidence is incomplete.  Cases (1) and (2) relies on several assumptions: that the capabilities tested are genuinely new, general, and don't rely on memorized training data or other shortcuts; that these capabilities are not present in simpler models; and that the capabilities are unexpected or unpredictable given the training data and the models' size.  None of these assumptions has been conclusively verified.  As for case (3) the complexity framework of \cite{Shalizi2025-yd} provides a principled approach to thinking about ``world models'' as these relate to discrete-time stochastic processes. To the extent that an LLM is effective at next-token prediction, and to the degree to which the model can be shown to exploit a minimum of information, they might be described as  world models. However, the recent work by \cite{vafa2024evaluating} demonstrates that recovering an accurate  world model is very difficult, since next token prediction is a fragile metric.

 LLMs display a wide range of functional capabilities, but due to their large number of parameters, nonlinear interactions among components, and size of training data, pinpointing which of these capabilities is ``surprising'' is incredibly difficult. Neural networks are known to be universal function approximators, and to the degree that many tasks, such as arithmetic and multiplication, are functions, perhaps LLMs have been implicitly ``programmed'' through their extensive training on language data.  
 
 There are three possible roles of language as it relates to training an LLM: (1) language itself provides a more or less complete and compressed representation of the world (including non-linguistic modalities); (2) spoken or written language mirrors an internal ``language of thought''; and (3) language is a non-supervised ``programming language''. If language does provide a complete representation of the world, then training on more language data would indeed enable an increasingly expansive and detailed representation of natural and cultural patterns and processes.  If natural language is the language of thought (``mentalese'') then training on more language data would fill out the numerous ways that humanity has historically reasoned about regularities in the world.  And if language is a programming language, by combining detailed instruction tuning with next word prediction it can exploit principles of computational universality to implement any computable function. 
 
 We do not have definitive evidence for any of these three claims, but they play a crucial role in any statement relating to how surprising the behavior of an LLM will be deemed. The more information about the world that resides in language, the weaker emergence claims become, under any of these claims. This is because in the limit of any KI system, the internal degrees of freedom of a model simply converge by some form of engineering on every external degree of freedom, and no coarse-grained internal model is produced or required. The fact that an internal program only works when complete (e.g., waiting for a final control loop to be implemented), and thus produces a discontinuous increase in performance for the system, does not in way make this increase an emergent property. 

The approach we adopted in this paper is to properly assess for emergence claims the properties of the structure responsible for a new capability, and to reserve the notion of emergence only when one or more the the conditions for emergence (Section~\ref{list}) are met. We do not yet have a theory with respect to capabilities themselves; we do not know a priori which tasks can be accomplished by LLMs as they just get larger (``more is more'') versus those that require a qualitative change in network internal structure (``more is different'').  But we believe that the approach and principles described in this paper will contribute to a more rigorous evaluation of emergence in complex AI models and a better scientific understanding of the extent to which scaling such models can produce increasingly sophisticated and general capabilities. 

\subsection{Emergent Capability Is Not Emergent Intelligence}

In this paper, we have situated claims of emergent LLM abilities in a broader framework for understanding emergence.  While most such claims focus on individual capabilities, some influential voices in the LLM community have gone further, asserting that LLMs are a form of ``emergent intelligence''.  Geoffrey Hinton has gone as far as stating that ``large scale digital computation is probably far better at acquiring
knowledge than biological computation and may soon be much more intelligent than us.'' \cite{hinton2024will}
 
In humans we do not mistake various, often impressive, cognitive capabilities for intelligence. Intelligence is reserved for a more general capacity for learning and solving problems quickly and parsimoniously, with a minimal expenditure of energy (a similar view is given in \cite{chollet2019measure}), where the concepts and methods used in solving particular problems can be easily modified and repurposed. It is for this reason that for our cognitively limited species, the hallmark of emergent intelligence consists in finding very compact and energetically efficient means of solving problems: ``less is more''. These solutions are moreover readily communicable with a few words of instruction between humans based on their understanding the meaning of what is communicated.  We do not teach one another mathematics by sharing synaptic weights, but by drawing a few illustrative figures, or manipulating symbols on a page. Indeed human cultural evolution can be understood as the transmission of more effective and compact symbols and procedures for inference \cite{mazur2014enlightening, rojas2025language}. 

This human form of intelligence can be considered truly emergent because it involves the discovery of effective theories of knowledge---compact and low-dimensional languages that screen-off most of neuroscience and even much of psychology. 
Indeed, folk psychology is in many ways a successful effective theory: its concepts  allow humans to attribute beliefs and desires to other people, and frequently successfully predict and explain their behavior \cite{Hutto21}. Movies, art, politics, and journals,  like the one in which this article appears, would not exist without shared folk psychological concepts.

The causal power of human understanding is also incontestable.  A recent paper illustrates this by contrasting how two different forms of intelligence---procedural and understanding-based, can  induce an identical neural network structure \cite{courellis2024abstract}.  Briefly, monkeys and humans can be trained to learn two sets of pairwise associations between arbitrary symbols. Each set of associations is defined as a context. What has been shown is that, through trial and error, monkeys and humans can generate a representational geometry across a large population of neurons in hippocampus that permits implicit inference of the latent contextual variable.  What is truly surprising is that by providing verbal instruction, a human (unlike a monkey) can configure this structure in minutes rather than through laborious training. This ``instant'' intelligence is based in this case on understanding an instructed sentence. An effective theory made of psychological constructs can causally change a neural network structure. There are many such examples of short circuiting of trial and error learning by quickly ``getting it’’.\cite{mazzoni2006implicit}\cite{ghilardi2009learning}\cite{manley2014money}. This includes comparisons across human subjects where only those individuals who explicitly understood a task (via a natural language explanation) reached a correct solution whereas implicit trial and error reinforcement failed to converge.  \cite{krakauer2022representation}.  This kind of instruction-mediated emergence has yet to be demonstrated in an LLM.

Human intelligence is a low-bandwidth phenomenon, and is as much if not more about the scaling down of effort as the scaling up of capability \cite{krakauer2010intelligent}. As Einstein wrote, ``The grand aim of all science is to cover the greatest number of empirical facts by logical deduction from the smallest number of hypotheses or axioms.'' \cite{barnett2005universe} We know that for any elegant algorithm there is an alternative brute force solution that does the job. It might even be the case that there are uncountable problems that require brute force and that this is a domain where LLMs and their cognitively alien relatives, including SAT solvers, will provide extraordinary utility \cite{heule2017science}.  What Donald Knuth said of programs might also be applied to intelligence: ``Programs are meant to be read by humans and only incidentally for computers to execute.'' \cite{abelson1996structure}. Similarly, intelligence is a property of understanding and only incidentally a matter of capability. 




\subsection{Acknowledgments}
The authors thank the referees for their thoughtful and constructive comments, which clarified many arguments in this paper. This work has been supported by the Templeton World Charity Foundation, Inc.\ (funder DOI 501100011730) under the grant https://doi.org/10.54224/20650. D.C.K. is also supported by grant no.\ 81366 from the Robert Wood Johnson Foundation on Using Emergent Engineering for integrating complex systems to achieve an equitable society. M.M. is also supported by subcontract 2673699 from the Sandia National Laboratories.

\bibliographystyle{unsrt}
\bibliography{emergence.bib}
\end{document}